%% file: main.tex
\documentclass[letterpaper]{article}
\usepackage{aaai}
\usepackage{times}
\usepackage{helvet}
\usepackage{courier}
\usepackage[all]{foreign}
\usepackage{bm}
\usepackage{amsmath}
\usepackage{xspace}
\usepackage{url}
\usepackage[inline]{enumitem}
\input{epistemicMacros}
\input{E-PDDL}
\input{MAR}
\usepackage{xcolor}

\newcommand{\pdkbpddl}{PDKB-PDDL\xspace}
\newcommand{\knobel}{knowledge\xspace}
\newcommand{\KnoBel}{Knowledge\xspace}
\newcommand{\rpmep}{RP-MEP\xspace}
\newcommand{\mypar}[1]{\medskip

\noindent\emph{#1}\hspace{\parindent}}
\newcommand{\mymail}[1]{\ensuremath{\mathtt{#1}}}

\usepackage{amsthm}
\theoremstyle{definition}
\newtheorem{example}{Example}

\begin{document}
%
\title{E-PDDL: A Standardized Way of Defining Epistemic Planning Problems}

\author{Francesco Fabiano\\DMIF Department\\ University of Udine, Italy \\ \mymail{francesco.fabiano@uniud.it}
\And Biplav Srivastava\\AI Institute \\ University of South Carolina, Columbia, USA \\ \mymail{biplav.s@sc.edu}
\And Marianna Bergamaschi Ganapini\\ Philosophy Department\\Union College, NY, USA \\ \mymail{bergamam@union.edu} 
\AND Jonathan Lenchner \and Lior Horesh \and Francesca Rossi\\Thomas J. Watson Research Center, Yorktown Heights, NY USA \\ \mymail{\{lenchner;lhoresh\}@us.ibm.com} \\\mymail{francesca.rossi2@ibm.com}
}

\maketitle
\begin{abstract}
\begin{quote}

\input{chapters/ch0_abstract}

\end{quote}
\end{abstract}

\input{chapters/ch1_motivation}
\input{chapters/ch2_background}
\input{chapters/ch3_contribution}
\input{chapters/ch4_parser}
\input{chapters/ch5_conclusion}


\bibliographystyle{aaai}
\bibliography{bibliography.bib}

\end{document}

%% file: epistemicMacros.tex
\usepackage{ifthen}
\newcommand*{\defemph}[1]{\ensuremath{\mathsf{#1}}}

\newcommand*{\fluent}[1]{\ensuremath{\defemph{#1}}}
\newcommand*{\agent}[1]{\ensuremath{\defemph{#1}}}

\newcommand{\agentSet}{\ensuremath{\mathcal{AG}}}

\newcommand{\B}[1]{\ensuremath{\textbf{B}_{\agent{#1}}}}

\newcommand{\C}[1]{\ensuremath{\textbf{C}_{\agent{#1}}}}

\newcommand*{\distract}[2]{%
	\ifstrequal{#2}{}%
	{\ensuremath{\mathtt{distract}(\agent{#1})}}%
	{\ensuremath{\mathtt{distract}(\agent{#1})\langle\agent{#2}\rangle}}%
}
\newcommand*{\open}[1]{%
	\ifstrequal{#1}{}%
	{\ensuremath{\mathtt{open}}}%
	{\ensuremath{\mathtt{open}\tuple{\agent{#1}}}}%
}
\newcommand*{\shout}[1]{%
	\ifstrequal{#1}{}%
	{\ensuremath{\mathtt{shout\_tails}}}%
	{\ensuremath{\mathtt{shout\_tails}\tuple{\agent{#1}}}}%
}
\newcommand*{\signal}[2]{%
	\ifstrequal{#2}{}%
	{\ensuremath{\mathtt{signal}(\agent{#1})}}%
	{\ensuremath{\mathtt{signal}(\agent{#1})\langle\agent{#2}\rangle}}%
}
\newcommand*{\peek}[1]{%
	\ifstrequal{#1}{}%
	{\ensuremath{\mathtt{peek}}}%
	{\ensuremath{\mathtt{peek}\tuple{\agent{#1}}}}%
}
\newcommand*{\tell}[2]{%
	\ifstrequal{#2}{}%
	{\ensuremath{\mathtt{tell}(\agent{#1})}}%
	{\ensuremath{\mathtt{tell}(\agent{#1})\langle\agent{#2}\rangle}}%
}
\newcommand*{\flip}[1]{%
	\ifstrequal{#1}{}%
	{\ensuremath{\mathtt{flip}}}%
	{\ensuremath{\mathtt{flip}\tuple{\agent{#1}}}}%
}

\newcommand*{\mAP}{\mbox{\ensuremath{m\mathcal{A}+}}}
\newcommand*{\mar}{\mbox{\ensuremath{m\mathcal{A}^\rho}}\xspace}

\newcommand*{\EFP}{\mbox{E{\footnotesize  FP}}\xspace}

\newcommand*{\sAG}{\ensuremath{\mathcal{AG}}}

\newcommand*{\shortimplies}{\ensuremath{\Rightarrow}}

\newcommand*{\bra}[1]{\ensuremath{\{#1\}}}
\newcommand*{\tuple}[1]{\ensuremath{\langle #1 \rangle}}



%% file: E-PDDL.tex
\usepackage{listings}
\lstdefinelanguage{e-PDDL}
{
  sensitive=false,    
  morecomment=[l]{;}, 
  alsoletter={:,-},   
  morekeywords={
    :agents,:observers,:p_observers,:executor,:act_type,diff,:depth,
    :derive,:effects,action:,:projection,:task,:init-type,
    define,domain,problem,not,and,or,when,forall,exists,either,
    :domain,:requirements,:types,:objects,:constants,
    :predicates,:action,:parameters,:precondition,:effect,
    :fluents,:primary-effect,:side-effect,:init,:goal,
    :strips,:adl,:equality,:typing,:conditional-effects,
    :negative-preconditions,:disjunctive-preconditions,
    :existential-preconditions,:universal-preconditions,:quantified-preconditions,
    :functions,assign,increase,decrease,scale-up,scale-down,
    :metric,minimize,maximize,
    :durative-actions,:duration-inequalities,:continuous-effects,
    :durative-action,:duration,:condition,:mep
  },
  keywordstyle = [2]{\sffamily\footnotesize},
  morekeywords = [2]{empty,ontic,epistemic,sensing,announcement},
}

%% file: MAR.tex
\lstdefinelanguage{MAR}
{
  sensitive=false,    
  alsoletter={:,-,_,;},   
  morekeywords={
    aware_of,fluent,action,if,executable,causes,observes,determines,announces,initially,goal,
    agent,
  },
}

%% file: chapters/ch0_abstract.tex
\emph{Epistemic Planning} (EP) refers to an automated planning setting where the agent reasons in the space of knowledge states and tries to find a plan to reach a desirable state from the current state. Its general form, the \emph{Multi-agent Epistemic Planning} (MEP) problem involves multiple agents who need to reason about both the state of the world and the information flow between agents.
In a MEP problem, multiple approaches have been developed recently with varying restrictions, such as considering only the concept of \emph{knowledge} while not allowing the idea of \emph{belief}, or not allowing for ``complex" modal operators such as those needed to handle \emph{dynamic common knowledge}. 
While the diversity of approaches has led to a deeper understanding of the problem space, the lack of a  standardized way to specify MEP problems independently of solution approaches has created difficulties in comparing performance of planners, identifying promising techniques, exploring new strategies like ensemble methods, and making it easy for new researchers to contribute to this research area.
To address the situation, we propose a unified way of specifying EP problems - the Epistemic Planning Domain Definition Language, E-PDDL. We show that E-PPDL can be supported by leading MEP planners and provide corresponding parser code that translates EP problems specified in E-PDDL into (M)EP problems that can be handled by several planners. 
This work is also useful in building more general epistemic planning environments where we envision a meta-cognitive module that takes a planning problem in E-PDDL, identifies and assesses some of its features, and autonomously decides which planner is the best one to solve it.

%% file: chapters/ch1_motivation.tex
\section{Motivation}

Multi-agent scenarios are ubiquitous in everyday life. We often need to make decisions based on information on the problem to be solved and on our knowledge, or belief, about the preferences and actions of other agents involved in, or impacted by, the decision.
It is therefore an epistemic reasoning task to make decisions in this scenarios.

Often such decision involve creating a plan, that is, a sequence of actions that, when executed, will lead to a desired goal state.
The research area of \emph{Multi-agent Epistemic Planning} (MEP) attempts to build artificial agents that can build such plans. 
MEP is a generalization of the concept of \emph{Epistemic Planning} (EP),
which refers to a single agent setting that reasons about knowledge and beliefs. In the MEP scenario, agents also reason about 
the information that other agents have of each other, and how such information flows between agents.

Several approaches to develop epistemic solvers/planners have been made across the community, see for example~\cite{baral2015action,muise2015planning,Fab20}.
While these solvers may have different restrictions, \eg considering only the concept of \emph{knowledge} while not allowing the idea of \emph{belief}, or not allowing for \emph{dynamic common \knobel}, all of them share the same goal: to plan while considering the information flows between the agents.

While the diversification of approaches in solving this problem certainly brings benefits to the entire community, with many specialized efficient planners that are suitable for specific scenarios, the lack of a standardized way to define input problems in multi-agent epistemic planning has led to the creation of ad-hoc languages that can become an obstacle 
when comparing, combining (i.e., ensemble), or testing different planners.
We believe that a unified way of expressing MEP problems would provide researchers with a faster and less error-prone way of defining standardized benchmarks and would also relieve them from the burden of having to learn a new language each time they need to make use of a different tool.
Moreover, it would allow for easy transfer of techniques and results between different planning scenarios and planner methodologies.

In this paper we therefore propose the Epistemic Planning Domain Definition Language (E-PDDL), a unified way of defining epistemic problems that, similarly to the well-known language PDDL for classical planning, could be regarded as the standard 
language to be used to express MEP problems.
Being able to define MEP problems in E-PPDL and then to map them into 
specific languages for existing planners provides also the necessary 
environment for a real-time choice of the best planner for a given MEP problem. Indeed, this work is part of a larger effort that aims to equip artificial agents with meta-cognitive capabilities that, given internal values, resources, and criteria, can identify some of the problem features and assess what skills are required to best solve it. In the context of this paper, this would mean to be able to select the most suitable planner for a MEP problem. This is greatly facilitated if there is a way to map the input problem specification into the language required by each specific planner.

In the rest of the paper, we  first give necessary background about EP and MEP with a running example. We then introduce the E-PDDL language and its key features, and we show how it encompasses current MEP formalisms. Next, we highlight implementation details in the parser and conclude.

%% file: chapters/ch2_background.tex
\section{Background}
Recently the concept of Artificial Intelligence has gained a lot of momentum both inside and outside the computer science research community. 
We are, in fact, witnessing an increasing number of autonomous tools that help users in tasks that range from the trivial and ordinary to the complex and very specialized.
In particular, the area of automated planning, that has always been widely employed especially at the industrial level, has been constantly pushed towards improvement.
Nowadays planners and schedulers are able to solve complex tasks in very short times while also accounting for factors such as incomplete knowledge, non-determinism, fault occurrences and so on \cite{helmert2006fast,lipovetzky2017best,richter2010lama}.

However, the main focus of these research efforts has been about reasoning within single agent domains. 
This type of setting mainly requires the agent to characterize her knowledge about the world and understand in which way she can manipulate it in order to achieve a particular goal.
But, while the single-agent setting can be satisfactory in many scenarios, multi-agent planning can be adopted to characterize a wider spectrum of real-world situations.
It is therefore natural that, thanks to the increasing interest in the area and the more powerful hardware support, the planning problem is now also broadly studied in the context of multiple agents.

\subsection{Multi-Agent Epistemic Planning}

Nevertheless, the presence of numerous agents in the planning setting poses a number of challenges: among which is to study the information flows that occurs between the agents. 
The area of planning that reasons in an environments where these streams of ``knowledge" need to be considered is known as \emph{Epistemic Multi-Agent Planning}, or simply MEP (see~\cite{fagin1994reasoning,bolander2017gentle} for a more detailed introduction to the topic).
MEP derives from the well-known research area of \emph{Dynamic Epistemic Logic} which uses logic to formalize and reason and reason on both the state of the world and on the \emph{information change} in dynamic domains. 
To be more concrete, as stated in~\cite{van2007dynamic}, ``\emph{information} is something that is relative to a subject who has a certain perspective on the world, called an \emph{agent}, and that is meaningful as a whole, not just loose bits and pieces. This makes us call it \emph{knowledge} and, to a lesser extent, \emph{belief}.'' 
In what follows, for brevity, we use the term ``\knobel'' to encapsulate \textit{both} the notions of an agent's knowledge and beliefs with respect to some information.
The concepts of knowledge and beliefs are captured by the same modal operator in Dynamic Epistemic Logic and their difference resides in structural properties that the epistemic states respect (see~\cite{fagin1994reasoning} for more details).
This means that thanks to epistemic planners we are able to reasons on domains where the actions of the agents affect, and are affected, by the state of the world and the \knobel of the agents (about both the world and the others' \knobel).

As already mentioned, MEP is concerned with information change and inherits concepts from Dynamic Epistemic Logic (DEL).
In particular, the language of well-formed DEL formulae used to express agents' \knobel, is defined as follows:
	\begin{equation*}
	    \varphi ::= \fluent{f} \mid \neg \varphi \mid \varphi \wedge \psi \mid \B{i} \varphi \mid \C{\alpha} \varphi,
	\end{equation*}
	\noindent where $\fluent{f}$ is a propositional atom called a \emph{fluent}, $ \agent{i} $ is an agent that belongs to the set of agents $\agentSet$ s.t. $|\agentSet| \geq 1$, $\varphi$ and $\psi$ are belief formulae and $ \emptyset \neq \alpha \subseteq \agentSet $.
	A \emph{fluent formula} is a DEL formula with no occurrences of modal operators.
    A \emph{belief formula} is recursively defined as follows:
	\begin{itemize}
		\item A fluent formula is a belief formula;
		\item If $\varphi$ is a belief formula and $\agent{i} \in \sAG$, then $\B{i}{\varphi}$ (``$ \agent{i} $ knows/believes that $ \varphi$'') is a belief
		      formula where the modal operator $\B{}$ captures the concept of \emph{\knobel};
		\item If $\varphi_1, \varphi_2$ and $\varphi_3$ are belief formulae,
		      then $\neg \varphi_3$ and $\varphi_1 \,\mathtt{op}\, \varphi_2$ are belief
		      formulae, where $\mathtt{op} \in \bra{\wedge,\vee, \shortimplies}$;
		\item If $\varphi$ is a belief formula and  $\emptyset \neq \alpha \subseteq \sAG$
		      then $\C{\alpha}\varphi$
		      is a belief formula where $\C{\alpha}$ captures the \emph{Common \knobel} of the set of agents $\alpha$.
		  \end{itemize}
    The formula $\C{\alpha}\varphi$ translates intuitively into the conjunction of the following belief formulae:
    \begin{itemize}
        \item every agent in $\alpha$ knows $\varphi$;
        \item every agent in $\alpha$ knows that every agent in $\alpha$ knows $\varphi$;
        \item and so on \emph{ad infinitum}.
    \end{itemize}

	The semantics of DEL formulae is traditionally expressed using \emph{pointed Kripke structures}~\cite{Kripke1963-KRISCO}. We refer the interested reader to~\cite{fagin1994reasoning,bolander2017gentle,Gerbrandy1997} for a comprehensive introduction to how Kripke structures or similar formalisms are used to capture the idea of an epistemic state and how the concept of entailment is defined.
	
	In this paper, we will make use of the \textbf{Coin in the Box} domain (Example~\ref{ex:coin_box}), first introduced in~\cite{baral2015action}, as a running example. In particular, we will present a three agent instance of this epistemic domain that will allow us to better illustrate some of the concepts we will later introduce.
	
	\begin{example}[\textbf{Coin in the Box}]\label{ex:coin_box}
	Three agents, \agent{A}, \agent{B}, and \agent{C}, are in a room where in the middle there
	is a box containing a coin. In the initial state it is known that:
	\begin{itemize}
		\item none of the agents know whether the coin lies heads or tails up;
		\item the box is locked and one needs a key to open it;
		\item some agents (usually only \agent{A}) have the key to the box;
		\item in order to learn whether the coin lies heads or tails up, an agent
		  can peek into the box, but this requires the box to be open; and
		\item agents can be either attentive or not towards the box.
	\end{itemize}
    Agents can perform the following actions, which have a straightforward meaning: 
	\begin{itemize}
		\item\emph{Opening} the box;
		\item\emph{Peeking} into the box;
		\item\emph{Signaling} another agent; 
		\item\emph{Distracting} another agent; and
		\item\emph{Announcing} the coin position.
	\end{itemize}
\end{example}
		
\subsection{Action Languages}

The field of planning has seen many representations. For example, in classical planning, there was STRIPS~\cite{strips}, Action Description Language (ADL)~\cite{adl} and SAS+~\cite{comparison-pl-backstorm} before Planning Domain Description Language (PDDL)~\cite{pddl,pddl2.1} standardized the notations. Nowadays, planners routinely use PDDL for problem specification even if they may convert to other representations later for solving  efficiency \cite{pddl-conversion-finite}.  PDDL envisages two files, a domain description file which specifies information independent of a problem like predicates and actions, and a problem description file which specifies the initial and goal states. PDDL provides a relaxed specification of the output plan --  it is a series of time steps, each of which can have one or more  instantiated actions with concurrency semantics. Specification of the correctness of plans was formalized much later when tools like VAL were created~\cite{plan-verif-val}.

In PDDL, a planning  environment is described in terms of objects in  the world, predicates that describe relations that hold between these objects, and actions that bring change to the world by manipulating relations. 
A problem is characterized by an initial state, together with a goal state that the agent wants to transition to, both states specified as configurations of objects.
When planning is used for epistemic reasoning, the objects in the problem can be physical (real world objects) as well as abstract (knowledge and beliefs).


\subsubsection{Epistemic Action Languages}
We briefly present the action languages adopted by two different epistemic solvers: \rpmep~\cite{muise2015planning} and \EFP~\cite{Fab20}.
We chose these solvers as, in is our opinion, they represent the state of the art epistemic planners. 
The former, \ie \rpmep, adopts a conversion to classical planning in order to tackle epistemic reasoning while the latter, \ie \EFP, comprehensively reasons on full-fledged epistemic states.
While the two planners are able to solve the same families of domains, important differences in their solving strategies make certain problem instances more suited to one approach rather than the other. For a detailed comparison of the solvers refer to~\cite{Fab20,le2018EFP}.
The main objective of this paper is to present a unified way of providing problem specification for the epistemic planning community so that the planners, starting from the two considered here, can be easily tested, confronted or even combined.
\mypar{\pdkbpddl} Let us start by presenting the main characteristics of the action language \pdkbpddl adopted by the solver \rpmep~\cite{muise2015planning,rpmep}.
This action language clearly stems from PDDL, the  de-facto standard for single-agent action languages, and allows one to express formulae that contain the modal operator $\B{}$.
While this language seems to represent a version of PDDL that is suitable for epistemic reasoning, 
there are some limitations that render \pdkbpddl not general enough for this purpose. 
In fact, while this language inherits the files schema from PDDL, also allowing for parametric action definitions, the epistemic components of the actions' description are tailored for the \rpmep solving process. This makes \pdkbpddl not flexible enough to be adopted as an input specification language for other planners (\eg \EFP). 
In particular, what makes this language not suitable for every epistemic approach is that, in order to have a valid \pdkbpddl domain, each action must specify the entirety of its effects explicitly.
While this is a common procedure in other types of reasoning in epistemic planning, the need for explicit effects specification brings some limitations.
To be more concrete, each action in \pdkbpddl needs to define not only how the action itself modifies the world but also how the \knobel of the agents are updated---that is, the belief update of each action needs to be explicit. 
This requirement is limiting because epistemic planners (\eg \EFP) may reason through implicit \knobel update\footnote{The distinction between explicit and implicit \knobel update derives from the fact that \rpmep grounds the problem while \EFP reasons directly on epistemic states.} and therefore ``automatically" derive how the the agents change their \knobel after an action execution.
While having explicit updates allows one to freely characterize each action's effects (\eg in Example~\ref{ex:coin_box} it would be possible to define a new action in which an agent opens the box but will believe that the coin is tails up), we feel that leaving such detail as input is partially in contrast with the idea of autonomous epistemic reasoning. 
Furthermore, it is easy for the user to inadvertently omit or mis-characterize ``intricate'' \knobel relations that should instead be handled by the autonomous solving process. For example, in writing the effects \emph{peeking} of Example~\ref{ex:coin_box} the effect ``the agent who peeked inside the box knows that the attentive agents know that he knows the coin position but the attentive agents themselves do not know the position of the coin" may be misrepresented.
This chain of \knobel is derived by partial observability (explained in the following paragraph) and, as well as other complex \knobel structures, may be easily mis-characterized given its complex and not intuitive nature.

%
%
%
\mypar{\mar} We can now introduce \mar, the language used to describe inputs for \EFP~\cite{Fab20}.
This language is an evolution of \mAP\ first introduced in~\cite{baral2015action}.
Unlike \pdkbpddl, \mar does not take inspiration from PDDL and introduces an ad-hoc syntax to define epistemic problems.
This syntax, illustrated in detail in~\cite{baral2015action,cilc19Epistemic}, describes an input problem through a single file that specifies initial and goal states (or rather the belief formulae that identify such states) along with the actions making each file a problem instance.
Moreover, unlike the input language of \rpmep, \mar defines only implicit effects and by associating each action with the agents' observability relations (which agent is aware of the action's occurrence and/or effects) and with a ``type" (\textit{world-altering}, \textit{sensing} and \textit{announcement}) is able to derive the complete \knobel update.
In particular, in~\cite{baral2015action}, each action execution \texttt{act} associates each agent  \agent{ag} to one of the following observibility relations:
\begin{itemize}
	\item \emph{fully observant} if \agent{ag} knows about the execution of \texttt{act} and about its effects on the world (\eg any attentive agent that witnessed the opening of the box in Example~\ref{ex:coin_box});
	\item \emph{partially observant} if \agent{ag} knows about the execution of \texttt{act} but she does not know how \texttt{act} affected the world (\eg any attentive agent that sees another agent \emph{peeking} inside the box in Example~\ref{ex:coin_box}); and
	\item \emph{oblivious} if \agent{ag} does not know about the execution of \texttt{act} (\eg any non-attentive agent during the execution of any action in Example~\ref{ex:coin_box}).
\end{itemize}
Moreover, as noted before, each action is associated with one of the following ``types":
\begin{itemize}
 \item \emph{World-altering} (or \emph{ontic}) action, used to modify certain properties (\ie fluents) of the world, \eg
	the action of \emph{opening} of Example~\ref{ex:coin_box}.
	\item \emph{Sensing} action: used by an agent to refine her beliefs about the world, \eg
	the action of \emph{peeking}.
	\item \emph{Announcement} action: used by an agent to affect the beliefs of other
	agents \eg in Example~\ref{ex:coin_box} the action of \emph{announcing}.
\end{itemize}
While this language is able to characterize implicit \knobel updates (from which the explicit effects can be derived), it lacks the structure of PDDL and relies on a custom syntax that may pose an obstacle to any researcher that is trying to approach the area of epistemic planning.
Moreover, another problem of \mar is the lack of parametric actions that, in several cases, forces the user to repeat the definition of the same action, with small variations, over and over.

%% file: chapters/ch3_contribution.tex
\section{E-PDDL}
\newcommand{\codetext}[1]{\ensuremath{\mathtt{#1}}}
\newcommand{\probdom}{problem-\emph{domain}\xspace}
\newcommand{\probinst}{problem-\emph{instance}\xspace}
\newcommand{\impltoexpl}{From Implicit to Explicit \KnoBel Update}

In what follows we present E-PDDL making use of the Coin in the Box domain (Example~\ref{ex:coin_box}) to better explain some of the features.
Let us remark that the syntax has been chosen with the objective of minimizing the difference with standard PDDL while providing a general epistemic input language.
We will start by showing the syntax of the \probdom---that contains the general settings of the problem---and then we will illustrate how a \probinst---that contains specific objects, initial conditions and goals---is characterized.
In particular in Listings~\ref{lst:domain} we present the characterization of the Example~\ref{ex:coin_box} \probdom and in Listings~\ref{lst:instance} we present a simple \probinst where it is known that agent \agent{a} has the key and the goal is for \agent{a} to know the coin position.
Before proceeding with the description of the E-PDDL syntax we need to introduce the meaning of the operator ``\codetext{[\agent{i}]}" where $\agent{i} \in \sAG$.
This operator captures the modal operator $\B{\agent{i}}$.
For example in Line 13 of Listings~\ref{lst:domain} the formula \codetext{[?\agent{i}](has\_key\ ?\agent{i})} reads ``agent \agent{i} knows \codetext{has\_key\_\agent{i}}" where $\agent{i} \in \sAG$ and the fluent \codetext{has\_key\_\agent{i}} encodes the fact that \agent{i} has the key.
When the operator is of the form ``[$\alpha$]" where $\alpha \subseteq \sAG$ and $|\alpha| \geq 2$ then it captures the idea of common \knobel. 

\subsection{Problem \emph{Domain}}
First of all let us note that in Listings~\ref{lst:domain} the actions \codetext{signal} and \codetext{distract} are omitted to avoid clutter. In fact, these actions are world-altering and, therefore, share a similar structure with the action \codetext{open}.

Following the PDDL syntax~\cite{pddl2.1} we start the \probdom definition by defining the name and the requirements of the problem (Lines 1 and 2-3 of Listings~\ref{lst:domain}, respectively). We included a new requirement called \codetext{:mep} to identify the need for E-PDDL.
In Lines 5-8 are introduced the predicates following the PDDL standard.
A small variation is the \emph{object-type} \codetext{agent} that does not need to be defined and is used to define variables that capture the acting agents.

From Line 10 to Line 19 the action \codetext{open} is introduced.
The action's definition starts with its \emph{name} (Line 10) and its \emph{type} (Line 11).
The concept of {action type} is inherited from \mar and for now is restricted to be one among ontic, sensing or announcement since these are the accepted variations of an action in the MEP community.
This concept allows to apply the right transition function in the context of implicit \knobel update.
We plan, for the future, to include custom action types through the definition of custom event models (see~\cite{baral2015action} for a better introduction on this topic).
Since such feature needs to be also implemented in the epistemic planners, we leave this contribution as a future work.
Next, in Line 12 and Line 13, respectively, the action's \emph{parameters} and \emph{preconditions} are defined.
The {parameters} have the same role that they have in PDDL, that is they are used to associate the variables of the action's definition with an \emph{object type}.
Similarly, also the field {preconditions}---identified by any belief formula---follow the standard PDDL meaning.
Following the {preconditions} the action specify the \emph{effects}.
The {effects}, contrarily from the {preconditions}, cannot contain arbitrary belief formulae.
In fact, depending on the action type, only fluents or special fluent formulae are allowed.
In particular, ontic actions allow for conjunctions of fluents while epistemic actions (\ie sensing and announcements) can only consider single fluents as effects.
These limitations, inherited by \mar, are necessary for planners that reason on epistemic transition functions.
We address the reader to~\cite{Fab20} for a more detailed description of the motivations behind these limitations.
Finally, the last field of the action \codetext{open} is about the \emph{observer}s.
This field is used to indicate which agents are fully observant, \ie knows about the execution and the effects of the action.
Knowing which agent is observant is once again useful to derive how the \knobel of the agents is updated after the action is been executed.
To better characterize the set of observant agents we introduced the operator \codetext{diff} that allow to ``isolate" the executor of the action (since the executor needs to be observant and should not depend on other factors).
For example, the condition in Lines 15-17 reads as: ``the agent \agent{i}, \ie the executor, is fully observant" (Line 15) and ``for every agent \agent{j} $\neq$ \agent{i} if \agent{j} \codetext{is\ looking} then \agent{j} is fully observant" (Lines 16-17).
The same schema is used to define partial observers, the ones that are aware of the action execution but do not know the results of such action, with the field \emph{p\_observers}; an example of partial observability is at Line 27.
\begin{lstlisting}[
  caption={E-PDDL Coin in the Box \probdom.},
  captionpos=b,
  breaklines=true,
  basicstyle=\footnotesize,
  numbers=left,
  stepnumber=1,
  postbreak=\mbox{{$\hookrightarrow$}\space},
  label={lst:domain},
  language=e-PDDL]
(define (domain coininthebox)
 (:requirements :strips 
  :negative-preconditions :mep)
  
 (:predicates (opened)
              (has_key ?i - agent)
              (looking ?i - agent)
              (tail))
 
 (:action open
   :act_type     ontic
   :parameters   (?i - agent)
   :precondition ([?i](has_key ?i))
   :effect       (opened)
   :observers    (and (?i)
                 (forall(diff(?j-agent)(?i))
                 (when(looking ?j) (?j))))
 )

 (:action peek
   :act_type	 sensing
   :parameters   (?i - agent)
   :precondition (and ([?i](opened))
                      ([?i](looking ?i)))
   :effect       (tail)
   :observers	 (?i)
   :p_observers  (forall(diff(?j-agent)(?i))
                 (when(looking ?j) (?j)))
 )

 (:action announce
   :act_type	 announcement
   :parameters   (?i - agent)
   :precondition ([?i](tail))
   :effect       (tail)
   :observers    (and (?i)
                 (forall(diff(?j-agent)(?i))
                 (when(looking ?j) (?j))))
 )
)
\end{lstlisting}
The introduced fields are tailored for implicit \knobel update but in Section ``\textbf{\impltoexpl}" we will explain how, with the presented syntax, it is possible to generate a valid input also for those planners that need the effects of the action to be completely explicit.

\subsection{Problem \emph{Instance}}
In Listings~\ref{lst:instance} we present an example of an E-PDDL \probinst.
In Line 1 and in Line 2 the \probinst name (\ie \codetext{toyinstance}) and the related \probdom name are defined, respectively.

Next, in Line 3, the \emph{object type} \codetext{agent} values are defined.
In this particular instance we defined three agents \agent{a}, \agent{b} and \agent{c} as in Example~\ref{ex:coin_box}.

Following, in Line 4, the \codetext{depth} is specified.
The concept of depth of a belief formula is used to identify the number of \emph{nested} epistemic operators.
For example, given two agents \agent{i} and \agent{j} the belief formula $\B{i}\varphi$ has depth 1 while $\B{i}\C{i,j}\varphi$  has depth 2.
This field is introduced to accommodate the need for certain planners, \eg \rpmep, to limit the depth of the belief formulae.
In fact \rpmep relies on grounding the formulae into classical planning ``facts" and without bound on these formulae could be infinite.
On the other hand, the limit on depth is ignored by the planners, \eg \EFP, that reason directly on epistemic states.

Lines 5-12 present the belief formulae that describe the initial state. The formulae are considered to be in conjunction with each other.
The \emph{initial conditions} only require to specify when a fluent is true and consider false whichever fluent is not specified (Line 5).
Moreover the initial conditions are also used to specify what is known in the initial state (Line 6-12).
While the belief formulae in this field can be of any type, let us remark that in~\cite{son2014finitary} it is demonstrated that to create a finite number of epistemic states from a set of formulae, this set must respect a finitary \textbf{S5} logic and therefore the \knobel must be expressed in terms of common \knobel.
This means that if the initial conditions do not comply with a finitary \textbf{S5} logic the planners that construct the initial epistemic state from the given specification may not work.

Finally in Line 13 the conjunction of belief formulae that represent the goals is defined.
Here, to respect Example~\ref{ex:coin_box}, only the goal that reads ``\agent{a} knows that the coin is tails up" is written.
\begin{lstlisting}[
  caption={E-PDDL Coin in the Box \probinst.},
  captionpos=b,
  breaklines=true,
  basicstyle=\footnotesize,
  numbers=left,
  stepnumber=1,
  postbreak=\mbox{{$\hookrightarrow$}\space},
  label={lst:instance},
  language=e-PDDL]
(define (problem toyinstance)
 (:domain coininthebox)
 (:agent a b c)
 (:depth 2)
 (:init (tails) (has_key a) (looking a)
        ([a b c](has_key a))
        ([a b c](not (has_key b)))
        ([a b c](not (has_key c))))
        ([a b c](not (opened)))
        ([a b c](looking a))
        ([a b c](not (looking b)))
        ([a b c](not (looking c))))
 (:goal ([a](tails)))
)\end{lstlisting}

\subsection{\impltoexpl}\label{subsec:expl}
Since we tailored E-PDDL syntax to represent actions with implicit \knobel update we need to explain how E-PDDL itself is a suitable language also for those planner, \eg \rpmep, that need the \knobel update to be explicit.
That is, we need a standard way of deriving the explicit agents' \knobel update from an E-PDDL action description.
While deriving explicit \knobel update is not a method that is embedded in the language itself in what follows we propose the strategy that we adopted in our implemented parser (presented in the following Section).

The strategy that we adopted in our parser is based on~\cite{baral2015action} transition function where the idea of agents' observability is used to derive consistent agents' \knobel about the actions effects and/or execution.
In what follows we present a \emph{\knobel derivation schema} that, starting from the agents' observability, is able to generate the explicit \knobel update related to a single action. These updates generate all the \knobel-chains of finite length $\ell$ where $\ell \in \{0,\dots,\mathtt{d}\}$ and  $\mathtt{d}$ is the value assigned to the field \codetext{:depth} in the \probinst (in Listings~\ref{lst:instance} $\mathtt{d} = 2$).
Namely we will have all the following chains:
\begin{itemize}
    \item a fully observant knows the action's effect ($\ell = 1$);
    \item a fully observant knows that another fully observant knows the action's effect ($\ell = 2$);
    \item a fully observant knows the chains of length 2 ($\ell = 3$);
    \item and so on until we have that $\ell = \mathtt{d}$.
\end{itemize}

\noindent Moreover, when partially observant agents are defined we also need to take into consideration their perspective on the \knobel update.
To do that we will need to automatically generate the following \knobel-chains (still limited by the given depth):
\begin{itemize}
    \item a partially observant knows that any chain of fully observant agents knows the action's effect ($\ell = 2$); and
    \item any chain, with $\ell \leq \mathtt{d}-2$, of fully/partially observant knows the chain of \knobel presented in the previous point.
\end{itemize}
In order to better integrate explicit \knobel update we decided to incorporate E-PDDL with an extra, non-mandatory field for the actions' specification.
This field, identified by \codetext{:exp\_effect} can be used to the identify the explicit \knobel update of the action by using an arbitrary belief formula.
Let us note that when this field is defined it will completely override the automatically derived \knobel update for the planners that make use of explicit \knobel update, \eg \rpmep.
On the other hand, planners that make use of a full fledged epistemic transition function, \eg \EFP, will ignore the \codetext{:exp\_effect} field.

%% file: chapters/ch4_parser.tex
\section{An E-PDDL Parser}
In this Section we will present the implementation of an E-PDDL parser.
This parser is able to take E-PDDL \probdom and \probinst files and automatically generate corresponding inputs for the solvers \rpmep and \EFP.
That is, the main objective of our parser is to generate a \pdkbpddl and a \mar description of the given input.
While this is just a first step of the integration of E-PDDL in the epistemic planning community, thanks to our parser  \rpmep and \EFP can already exploit the newly presented syntax.
Moreover, this parser allows to better compare and/or even combine the two approaches.
Therefore, the presented approach may help in building a more general epistemic planning environment where, thanks also to a meta-cognitive reasoning module, takes a planning problem in E-PDDL, identifies and assesses some of its features, and autonomously decides which planner is the best one to solve it.

The parser source code, along with some examples and a small guide on how to use the code itself, is available at the following url: \url{https://github.com/FrancescoFabiano/E-PDDL}.
Our E-PDDL parser is written in Python3~\cite{10.5555/1593511} and inherits its basic structure from an open source PDDL parser~\cite{pddlparser}.

As already said the objective of the parser is to ``transform" an E-PDDL input into files that can be directly used as input for either \rpmep or \EFP.
The idea is to provide the researchers with a tool that generates an output that can be directly used by the two above-mentioned planners. 

While elaborating upon the details of the code will be tedious and not informative we decide to show the result of the parsing process.
In Listings~\ref{lst:rpmepdom} and~\ref{lst:rpmepinst} show the \probdom and the \probinst (partial) input files for the planner \rpmep resulted by calling the parser on the \textbf{Coin in the Box} instance described in Listings~\ref{lst:domain} and~\ref{lst:instance}.
Listings~\ref{lst:efp}, on the other hand, presents (part of) the encoding of the same problem for the planner \mar.
Let us remark that Listings~\ref{lst:rpmepdom} and~\ref{lst:rpmepinst} follow the syntax of \pdkbpddl while Listings~\ref{lst:efp} respects the syntax of \mar.

\begin{lstlisting}[
  caption={\pdkbpddl Coin in the Box \probdom generated through the parsing process of Listings~\ref{lst:domain}.},
  captionpos=b,
  breaklines=true,
  basicstyle=\footnotesize,
  numbers=left,
  stepnumber=1,
  postbreak=\mbox{{$\hookrightarrow$}\space},
  label={lst:rpmepdom},
  language=e-PDDL]
(define (domain coininthebox)
 (:agents a b c)
 (:constants)
 (:types)
 (:predicates (opened) (has_key ?i - agent)
              (looking ?i - agent) (tails))

 (action: open
  :parameters	(?i - agent)
  :derive       (looking $agent$)
  :precondition ([?i](has_key ?i))
  :effects      (and (opened))
 )

 (action: peek
  :parameters	(?i - agent)
  :derive       (never)
  :precondition (and ([?i](opened))
                     ([?i](looking ?i)))
  :effects      (and [?i](tails)
                (forall (?j - agent)
                (when (looking ?j)
                [?j][?i](or(tails)(!tails)))))
 )

 (action: announce
  :parameters	(?i - agent)
  :derive       (looking $agent$)
  :precondition ([?i](tails))
  :effects      (and (tails))
 )
)
\end{lstlisting}

In Lines 10 and 28 of Listings~\ref{lst:rpmepdom} we can see how, thanks to the \pdkbpddl field \codetext{:derive-condition} (shorten to \codetext{:derive}), we are able to define which agents are aware of an action's execution and its effects. This special field automatically expand the \knobel update for \rpmep.
This conversion (that only allows belief formulae of depth up to 2 as indicated by Line 4 of Listings~\ref{lst:instance}) correctly captures the idea that fully observant agents are aware of the actions' effects and also know that the other fully observants know these effects.

On the other hand, to express the complete \knobel update of the action \codetext{peek} we need to account also for partial observability.
In Lines 20-23 the effects of the action are written so that the acting agent, \ie agent \agent{i}, knows that the coin is tails up.
The other agents, which are partially observant, only knows that \agent{i} knows the coin position but they do not the position themselves.
Given that the max depth of the formulae is set to 2 these effects suffice in capturing the behavior of \codetext{peek}.
Let us remark that the syntax presented in Lines 20-23 of Listings~\ref{lst:rpmepdom} is not fully compatible with \pdkbpddl but nicely captures the idea we are trying to express.
Partial observability, while supported by the parser, is not yet fully supported during the conversion to \pdkbpddl and is currently under development.

\begin{lstlisting}[
  caption={\pdkbpddl Coin in the Box \probinst generated through the parsing process of Listings~\ref{lst:instance}.},
  captionpos=b,
  breaklines=true,
  basicstyle=\footnotesize,
  numbers=left,
  stepnumber=1,
  postbreak=\mbox{{$\hookrightarrow$}\space},
  label={lst:rpmepinst},
  language=e-PDDL]
(define (problem pb1)
 (:domain coininthebox)
 (:depth 2)
 (:projection )
 (:task valid_generation)
 (:init-type complete)
 
 (:init
  (tails)
  (has_key a)
  (looking a)

  ;Length 1
  (forall ?ag1 - agent
	[?ag1](has_key a))

  ;Length 2
  (forall ?ag1 - agent
    (forall ?ag2 - agent
      [?ag2][?ag1](has_key a)))

 )

 (:goal 
  [a](tails)
 )
)
\end{lstlisting}
While Listings~\ref{lst:rpmepinst} has some extra fields w.r.t. to Listings~\ref{lst:instance} that are used by \rpmep to set some internal parameters the only notable difference is the conversion of the common \knobel operator, not supported by \rpmep, into chain of belief formulae.
These chains, bounded by the depth limit (that is our case is set to two), are the chains of \knobel that are used to define the common \knobel operator.
The conversion, if applied for example to Line 6 of Listings~\ref{lst:instance} ``\codetext{[\agent{a\ b\ c}](has\_key\ \agent{a})}", generates all the belief formulae where:
    \begin{enumerate*}[label= ]
        \item every agent in $\{\agent{a};\agent{b};\agent{c}\}$ knows \codetext{has\_key\ \agent{a}} ($\ell = 1$); and
        \item every agent in  $\{\agent{a};\agent{b};\agent{c}\}$ knows that every agent in  $\{\agent{a};\agent{b};\agent{c}\}$ knows \codetext{has\_key\ \agent{a}} ($\ell = 2$).
    \end{enumerate*}
    
\begin{minipage}{1.0\linewidth}
\begin{lstlisting}[
  caption={\mar Coin in the Box problem generated through the parsing process of Listings~\ref{lst:domain} and~\ref{lst:instance}.},
  captionpos=b,
  breaklines=true,
  basicstyle=\footnotesize,
  numbers=left,
  stepnumber=1,
  postbreak=\mbox{{$\hookrightarrow$}\space},
  label={lst:efp},
  language=MAR]
fluent has_key_a, has_key_b, .., tails;

action open_a, open_b, .., announce_c;

agent a, b, c;

executable open_a if B(a,has_key_a);
open_a causes opened;
b observes open_a if look_b;
c observes open_a if look_c;
a observes open_a;

executable peek_a if B(a,opened),B(a,look_a)
peek_a determines tails if look_a;
a observes peek_a;
b aware_of peek_a if look_b;
c aware_of peek_a if look_c;

executable announce_a if B(a,tails);
announce_a announces tails;
b observes announce_a if look_b;
c observes announce_a if look_c;
a observes announce_a;

initially tails, has_key_a, look_a;
initially -look_b, -look_c, ...;

initially C([a,b,c],look_a);
initially C([a,b,c],has_key_a);
initially C([a,b,c],-opened);
...
initially C([a,b,c],-has_key_c);

goal B(a,tails);
\end{lstlisting}
\end{minipage}

Finally, in Listings~\ref{lst:efp}, we show a partial results of the conversion of E-PDDL into \mar.
Since \mar does not allow to use variables and types to define the problem specification the parser grounds the instance of E-PDDL and use the result to generate the input for \EFP.
In particular, this grounded instance contains grounded fluents, \eg \codetext{has\_key\_a}, \codetext{has\_key\_b} and \codetext{has\_key\_c}, and grounded actions, \eg \codetext{open\_a}. 
The grounded actions are instantiated versions of the ones declared in E-PDDL where each variable is associated with one of its possible values.
Other small syntactical changes are then applied, \eg ontic actions are associated with keyword \codetext{causes} and partial observability is identified by the keyword \codetext{aware\_of}.

%% file: chapters/ch5_conclusion.tex
\section{Conclusion}

In this paper, we proposed E-PDDL, an Epistemic Planning Domain Definition Language, that represents a unified way to specify \emph{Multi-agent Epistemic Planning} (MEP) problems.
We show that E-PDDL can unify and support leading MEP planners, and provide a parser code to the community that is able to generate input specification for two state of the art epistemic reasoners.
Defining a unified language is the first step towards defining common benchmarks and evaluation metrics for all the MEP community.
Moreover, since the new language follows the well-know language PDDL, E-PDDL may also serve as a gateway for researchers already involved in the area of planing that want to investigate the branch of epistemic reasoning.
Finally defining a unified way of specifying the input for different epistemic solvers can help in designing a general purpose epistemic solver that is automatically able to select the best solving strategy, between existing planners, starting from the problem definition that can now be  ``understood" by every considered approach.
We leave the design of such architecture as future work.

While E-PDDL is able to capture most of the features that can be found in an epistemic domain the language only considers three types of actions: ontic, sensing and announcement.
We plan to expand the language syntax so that is able to accept custom action definitions, specified with a custom event model.
Nonetheless, before introducing such feature we must ensure compatibility with the considered epistemic solvers.
For \rpmep the conversion to explicit \knobel update is handled directly by the parser and, therefore, custom actions definition would only generate different chains of belief formulae.
On the other hand, \EFP relies on an epistemic transition function that expects the actions to be ontic, sensing or announcement and it would not be able to handle custom action types.
That is why before expanding E-PDDL we need to implement in \EFP the possibility of using custom action definitions. 

Finally we plan to refine the parser code to better integrate the concept of partial observability and to be more flexible in the definition of the formulae.